\documentclass{article}
\usepackage{spconf,amsmath,graphicx}
\usepackage{bm}
\usepackage{algorithm}
\usepackage{algpseudocode}
\usepackage{booktabs}
\usepackage{pifont}
\title{NFCDS: A PLUG-AND-PLAY NOISE FREQUENCY-Controlled DIFFUSION Sampling Strategy FOR IMAGE RESTORATION}
\name{Zhen Wang, \qquad Hongyi Liu*\thanks{This work was supported in part by the National Natural Science Foundation of China under Grant 61971223, in part by the Key Laboratory of Analysis of Mathematical Theory and Modeling of Complex Systems, the Ministry of Industry and Information Technology. Corresponding author:Hongyi Liu.}, \qquad Jianing Li, \qquad Zhihui Wei}

\address{Nanjing University of Science and Technology, Nanjing, China}

\begin{document}
%\ninept
%
\maketitle
\begin{abstract}
Diffusion sampling-based Plug-and-Play (PnP) methods produce images with high perceptual quality but often suffer from reduced data fidelity, primarily due to the noise introduced during reverse diffusion. To address this trade-off, we propose Noise Frequency-Controlled Diffusion Sampling (NFCDS), a spectral modulation mechanism for reverse diffusion noise. We show that the fidelity–perception conflict can be fundamentally understood through noise frequency: low-frequency components induce blur and degrade fidelity, while high-frequency components drive detail generation. Based on this insight, we design a Fourier-domain filter that progressively suppresses low-frequency noise and preserves high-frequency content. This controlled refinement injects a data-consistency prior directly into sampling, enabling fast convergence to results that are both high-fidelity and perceptually convincing—without additional training. As a PnP module, NFCDS seamlessly integrates into existing diffusion-based restoration frameworks and improves the fidelity–perception balance across diverse zero-shot tasks.
\end{abstract}
\begin{keywords}
Diffusion Model, Image Restoration, Plug-and-Play, NFCDS
\end{keywords}
\section{Introduction}\label{sec1}

Image restoration plays a crucial role in numerous practical applications,  aiming to reconstruct a high-quality original image $\bm{x}$ from degraded measurements $\bm{y}$. The degradation process is commonly modeled as:
\begin{align}
    \bm{y} = \bm{A}\bm{x}+\bm{n},
    \label{eq1}
\end{align}
where $\bm{A}$ denotes the degradation operator (e.g., a blur kernel or a downsampling matrix), which is typically non-invertible or ill-conditioned, and $\bm{n}\sim\mathcal{N}(\bm{0},\sigma_{\bm{y}}^2\bm{I})$
represents additive white Gaussian noise. Due to the inherent non-uniqueness of solutions and high sensitivity to noise, effective image restoration necessitates the incorporation of suitable prior knowledge to regularize the solution and ensure both stability and physical plausibility.

% Traditional model-based methods employ handcrafted priors (e.g., total variation(TV)\cite{tv} or sparsity\cite{madarkar2021sparse}), which offer theoretical guarantees but often lack the expressive capacity to capture complex image statistics, leading to suboptimal restoration performance. In contrast, deep learning methods achieve powerful performance via data-driven learning\cite{dl1,dl2}, however, due to high training costs, dependence on paired data, and poor generalization, their applicability to real-world scenarios is limited.
Traditional model-based methods use handcrafted priors (e.g., total variation (TV)\cite{tv} or sparsity\cite{madarkar2021sparse}), which offer theoretical guarantees but often lack the expressiveness to capture complex image statistics, leading to suboptimal restoration. In contrast, deep learning methods achieve strong performance through data-driven learning\cite{dl1,dl2}, however, their real-world applicability is limited by high training costs, reliance on paired data, and poor generalization.

% Zero-shot image restoration has garnered increasing attention, particularly through Plug-and-Play (PnP) methods\cite{plug1,plug2}. These methods embed pre-trained denoisers into iterative optimization, achieving strong performance without task-specific training. Diffusion models\cite{ddpm,ncsn,score} provide a powerful generative prior for high-fidelity image synthesis and have been successfully applied to PnP-based restoration\cite{diffpir,ddpg,ddlg}, significantly improving restoration quality.
Zero-shot image restoration has gained increasing attention, especially via Plug-and-Play (PnP) methods\cite{plug1,plug2}. These methods embed pre-trained denoisers into iterative optimization to deliver powerful results without task-specific training. Diffusion models\cite{ddpm,ncsn,score} provide a powerful generative prior for high-fidelity synthesis and have been successfully integrated into PnP-based restoration\cite{diffpir,ddpg,ddlg}, significantly enhancing quality.

% Current diffusion sampling-based PnP methods incorporate data consistency constraints into the reverse diffusion process and re-inject noise at each step to steer restoration along stochastic trajectories. While this mechanism enhances texture realism, it often compromises data fidelity—highlighting the dual role of noise: promoting fine detail synthesis while degrading structural accuracy.
Current diffusion sampling-based PnP methods enforce data consistency in the reverse diffusion process and re-inject noise at each step to guide restoration along stochastic trajectories. While this enhances texture realism, it often compromises data fidelity—revealing noise’s dual role: aiding detail synthesis while degrading structural accuracy.

% Existing works that improve upon noise focuses on forward-process noise scheduling (e.g., cosine in IDDPM\cite{IDDPM} or SNR-adaptive in Simple Diffusion\cite{simple}) or sampling trajectory optimization\cite{ding2025rass}—mostly for unconditional generation—and assumes spectrally uniform noise, ignoring frequency-dependent effects in restoration. Therefore, we revisit sampling noise from a frequency-domain perspective and propose a simple, training-free modulation strategy that jointly improves fidelity and perceptual quality.
Therefore, some works have sought to improve restoration by scheduling noise. However, these methods mainly address forward-process noise scheduling (e.g., cosine in IDDPM\cite{IDDPM} or SNR-adaptive in Simple Diffusion\cite{simple}) or sampling trajectory optimization\cite{ding2025rass}, primarily for unconditional generation, and assume spectrally uniform noise, overlooking frequency-dependent effects critical to image restoration.

\begin{itemize} 
\item We provide the first frequency-domain analysis of injected noise in diffusion sampling, showing its dual role in restoration: Low-frequency noise degrades data-consistent global structures, while high-frequency noise enhances realistic texture synthesis.
\item We propose a Noise Frequency-Controlled Diffusion Sampling (NFCDS) strategy that suppresses low-frequency components in the injected noise via a soft-thresholding mask, effectively mitigating structural drift caused by redundant perturbations.
\item Experiments on super-resolution and denoising across a benchmark dataset show that our method, without any additional training, can be directly integrated into existing sampling pipelines and achieves a superior balance between reconstruction fidelity and perceptual quality.
\end{itemize}

\section{Background}\label{sec2}
\subsection{Denoising Diffusion Probabilistic Model}\label{sec2.1}
Denoising Diffusion Probabilistic Models (DDPM\cite{ddpm}) generate samples by reversing a forward diffusion process that gradually adds Gaussian noise to a clean image until it becomes a standard Gaussian. This process is a fixed Markov chain over $T$ timesteps. Due to Gaussian distribution properties, the noisy sample at any timestep $t$ admits a closed-form expression directly from the original image $\bm{x}_0 \sim p_{\text{data}}$:
% Denoising Diffusion Probabilistic Models (DDPM\cite{ddpm}) generate samples by reversing a forward diffusion process that gradually transforms a clean image into a known prior distribution, typically standard Gaussian noise. This forward process is a fixed Markov chain defined over $T$ timesteps. Owing to the properties of Gaussian distributions, the noisy sample at any timestep $t$ can be expressed in closed form directly from the original clean image $\bm{x}_0 \sim p_{\text{data}}$:
\begin{align}
    \bm{x}_t = \sqrt{\bar{\alpha}_t} \, \bm{x}_0 + \sqrt{1 - \bar{\alpha}_t} \, \bm{\epsilon},
    \label{eq2}
\end{align}
where $\bm{\epsilon} \sim \mathcal{N}(\bm{0}, \bm{I}), \alpha_t = 1 - \beta_t$, $\bar{\alpha}_t = \prod_{i=1}^t \alpha_i$ and $\{\beta_t\}_{t=1}^T$ is a pre-defined variance schedule satisfying $0 < \beta_t < 1$.

During sampling, the reverse process reconstructs data by iteratively denoising from pure noise $\bm{x}_T \sim \mathcal{N}(\bm{0}, \bm{I})$. DDPM parameterizes this reverse dynamics using a neural network $\bm{\epsilon}_\theta(\bm{x}_t, t)$ to predict the noise added at step $t$. The update rule for one reverse step in DDPM is:
\begin{align}
    \bm{x}_{t-1} = \frac{1}{\sqrt{\alpha_t}} \left( \bm{x}_t - \frac{\beta_t}{\sqrt{1 - \bar{\alpha}_t}} \, \bm{\epsilon}_\theta(\bm{x}_t, t) \right) + \sqrt{\beta_t} \bm{\epsilon}_t,
    \label{eq3}
\end{align}

% DDPM requires both forward and reverse processes to be Markovian Gaussian chains, typically demanding hundreds to thousands of sampling steps. In contrast, DDIM shows that as long as the noise prediction objective matches that used during training, the reverse process need not preserve the original Markovian structure, thereby enabling more flexible generative trajectories:
Further, DDIM\cite{ddim}, as a faster generation scheme, offers more flexible sampling trajectories:
\begin{align}
    \bm{x}_{t-1} = \sqrt{\bar{\alpha}_{t-1}} \bm{x}_{0|t} + \bar{\sigma}_t \, \bm{\epsilon}_\theta(\bm{x}_t, t) + \sigma_t\bm{\epsilon}_t,
    \label{eq4}
\end{align}
where $\bm{\epsilon}_t\sim \mathcal{N}(\bm{0},\bm{I})$ and
\begin{align}
    &\bar{\sigma}_t= \sqrt{1 - \bar{\alpha}_{t-1}-\sigma_t^2}, \label{eq5}\\
   & \bm{x}_{0|t}=\frac{\bm{x}_t - \sqrt{1 - \bar{\alpha}_t} \, \bm{\epsilon}_\theta(\bm{x}_t, t)}{\sqrt{\bar{\alpha}_t}}.
    \label{eq6}
\end{align}
$\bm{x}_{0|t}$ denotes the estimate of the original image $\bm{x}_0$ given the noisy sample $\bm{x}_t$, which essentially equates to a single Gaussian denoising step. 

Thus, DDIM sampling can be viewed as an alternating denoising and re-noising process, and $\sigma_t$ controls the trade-off between predicted and random noise at sampling steps.

\subsection{Restoration via Plug-and-Play Diffusion Models}\label{sec2.2}

Traditional model-based methods employ a variational framework that constrains the solution space via a prior-based regularization term on $\bm{x}$, with the following simplified form:
\begin{align}
    \min_{\bm{x}} L(\bm{x}) = \underbrace{\ell(\bm{x},\bm{y})}_\text{data fidelity term} + \underbrace{s(\bm{x})}_{\text{prior term}}.
    \label{eq7}
\end{align}

The PnP framework offers a flexible approach to image restoration. Unlike traditional model-based methods that design explicit regularizers, PnP decouples data fidelity from prior modeling by plugging in an external denoiser as an implicit prior. Specifically, he framework regards the regularizer $s(\cdot)$ as a strong denoising operator $D(\cdot,\sigma_t)$, alternating data consistency and denoising per iteration. The proximal gradient descent update rule is followed:
% The PnP framework is a flexible method for solving imaging inverse problems. Unlike traditional model-based methods that require designing regularization terms, PnP decouples data fidelity from prior modeling by using an external denoiser as an implicit regularizer. It replaces the explicit regularizer $s(\cdot)$ with a strong prior-driven denoising operator $D(\cdot,\sigma_t)$, and implements it through iterative strategies  to perform data consistency updates and denoising at each step, progressively converging toward the optimal solution. The update rule based on proximal gradient descent is given as follows:
\begin{align}
    \left\{
    \begin{aligned}
    &\bm{x}_{0|t}=\mathcal{D}(\bm{x}_{t},\sigma_{t})\\
    &\bm{x}_{t-1}=\bm{x}_{0|t}-\mu_{t}\nabla_{\bm{x}}\ell(\bm{x}_{0|t},\bm{y})
    \end{aligned}
    \right..
    \label{eq8}
\end{align}
Here, $\mu_t$ is the step-size, and $\nabla_{\bm{x}}\ell(\bm{x}_{0|t},\bm{y})$ enforces the data consistency constraint. This mechanism avoids explicit prior modeling and task-specific retraining, achieving high restoration quality with drastically reduced training overhead.

% Diffusion models are multi-scale iterative denoisers that learn natural image priors during training. 
Sampling-based PnP integrates PnP into diffusion sampling by aligning each update with the model’s denoising trajectory, preserving data consistency while following a high-quality generative path for more realistic reconstruction. Its typical update rule is:
\begin{align}
    \left\{
    \begin{aligned}
    &\bm{x}_{0|t}=\mathcal{D}(\bm{x}_{t},\sigma_{t})=\frac{1}{\sqrt{\bar{\alpha}_t}}(\bm{x}_t - \sqrt{1 - \bar{\alpha}_t} \, \bm{\epsilon}_\theta(\bm{x}_t, t))\\
    &\bm{x}_{t-1}=\sqrt{\bar{\alpha}_{t-1}}\bm{x}_{0|t}-\mu_{t}\nabla_{\bm{x}}\ell(\bm{x}_{0|t},\bm{y})+ \bar{\sigma}_t \, \bm{\epsilon}_\theta(\bm{x}_t, t) + \sigma_t\bm{\epsilon}_t
    \end{aligned}
    \right.,
    \label{eq9}
\end{align}
where $\nabla_{\bm{x}}\ell(\bm{x}_{0|t},\bm{y})$ is the data-fidelity guidance and $\mu_t$ is the guidance scaling factor. 

It can be observed that the core of sampling-based PnP variants lies in injecting noise into the traditional iterative denoising framework to introduce controllable stochasticity and thereby achieve more realistic restoration. 

Current mainstream methods, such as DD-NRLG\cite{ddlg} and DiffPIR\cite{diffpir}, typically adopt this strategy, differing primarily in the formulation of their data consistency constraints. For instance, in DD-NRLG, the data consistency term is given by:
\begin{align}
 \nabla_{\bm{x}}\ell&(\bm{x}_{0|t},\bm{y})\simeq \nonumber\\
    &-\frac{1}{\sqrt{\bar{\alpha}_{t}}} \bm{A}^{T} \left( \frac{1-\bar{\alpha}_{t}}{\bar{\alpha}_{t}} \bm{A} \bm{A}^{T} + \sigma_{\bm{y}}^{2} \bm{I} \right)^{-1} 
    \cdot \left( \bm{y} - \bm{A}\bm{x}_{0|t} \right).
    \label{eq10}
\end{align}

% However, despite their improved visual quality, these methods exhibit notably lower data fidelity compared to traditional iterative denoising approaches—a degradation that can be attributed to the stochasticity introduced during the sampling process. To address this, this paper aims to thoroughly analyze the mechanism of noise injection, identify the key factors governing the trade-off between data fidelity and perceptual quality, and thereby enhance the data consistency performance of existing sampling-based methods.
% However, despite their improved visual quality, these methods exhibit significantly lower data fidelity than traditional iterative denoising approaches. This limitation stems from the stochasticity introduced during sampling, which is precisely the focus of our study.
Despite their superior visual quality, these methods suffer from substantially lower data fidelity than traditional iterative denoising approaches—a limitation rooted in the sampling-induced stochasticity that our study specifically addresses.

\section{Method}\label{sec3}
% In this section, to elucidate the trade-off between perception and fidelity induced by injected noise, we present a frequency-domain analysis of its role in both generative modeling and DD-NRLG-based image restoration.
To clarify the perception–fidelity trade-off caused by injected noise, we present a frequency-domain analysis of its role in image generation and DD-NRLG-based image restoration.

Images exhibit a distinct frequency-domain energy distribution: low frequencies dominate smooth regions and global structure, while high frequencies capture fine details such as edges and textures. Any signal $\bm{I}$ can be decomposed into its low- and high-frequency components:
% Images exhibit a distinct energy distribution in the frequency domain: low-frequency components dominate smooth regions and global structure, while high-frequency components correspond to fine details such as edges and textures. Any signal $\bm{I}$ can be uniquely decomposed into the sum of its low-frequency and high-frequency components:
\begin{align}
    \bm{I} = \bm{I}^{\text{LF}} + \bm{I}^{\text{HF}}.
    \label{eq11}
\end{align}
This decomposition can be achieved using frequency-domain transforms such as the Fourier transform.

\subsection{Role of Injected Noise in Image Generation}\label{sec3.1}
In image generation, the clean image $\bm{x}_0$ is completely unknown, and the generative process is entirely noise-driven. The reverse update formula of DDIM is:
\begin{align}
    \bm{x}_{t-1} = \sqrt{\bar{\alpha}_{t-1}} \, {\bm{x}}_{0|t} + \sqrt{1 - \bar{\alpha}_{t-1}} \, \bar{\bm{\epsilon}}_{t-1},
    \label{eq12}
\end{align}
where ${\bm{x}}_{0|t} =\frac{\bm{x}_t - \sqrt{1 - \bar{\alpha}_t} \, \bm{\epsilon}_\theta(\bm{x}_t, t)}{\sqrt{\bar{\alpha}_t}}$ is the current estimate of $\bm{x}_0$, and $\bar{\bm{\epsilon}}_{t-1} = \sqrt{1-\zeta}\,\bm{\epsilon}_\theta(\bm{x}_t, t)+\sqrt{\zeta}\,\bm{\epsilon}_{t}$ is a weighted combination of $\bm{\epsilon}_\theta(\bm{x}_t, t)$ and $\bm{\epsilon}_{t}$. This is a common variant of the DDIM sampling rule (Eq.(\ref{eq4})) that maintains generation quality while controlling diversity via the parameter $\zeta \in [0,1]$ .

By applying frequency-domain decomposition to Eq.(\ref{eq12}), we obtain the decoupled form as follows:
\begin{align}
    \bm{x}_{t-1} &= \underbrace{(\sqrt{\bar{\alpha}_{t-1}} \, {\bm{x}}_{0|t}^{LF} + \sqrt{1 - \bar{\alpha}_{t-1}} \, \bar{\bm{\epsilon}}_{t-1}^{LF})}_{\bm{x}_{t-1}^{LF}}\nonumber\\&+\underbrace{(\sqrt{\bar{\alpha}_{t-1}} \, {\bm{x}}_{0|t}^{HF} + \sqrt{1 - \bar{\alpha}_{t-1}} \, \bar{\bm{\epsilon}}_{t-1}^{HF})}_{\bm{x}_{t-1}^{HF}}.
    \label{eq13}
\end{align}

This formulation explicitly decomposes each sampling step into independent low- and high-frequency updates. Meanwhile, the noise $\bar{\bm{\epsilon}}_{t-1}$ is also split into its low-frequency $\bar{\bm{\epsilon}}_{t-1}^{LF}$ and high-frequency $\bar{\bm{\epsilon}}_{t-1}^{HF}$ components, enabling a detailed analysis of each frequency component’s role in the generative process.

% Thus, each sampling step is decomposed into updates of high- and low-frequency components: the original full-band Gaussian noise $\bm{\bar{\epsilon}}_{t-1}$ is split into its low-frequency part $\bm{\bar{\epsilon}}_{t-1}^{LF}$ and high-frequency part $\bm{\bar{\epsilon}}_{t-1}^{HF}$. 
\begin{figure}[h]
    \centering
    \includegraphics[width=\linewidth]{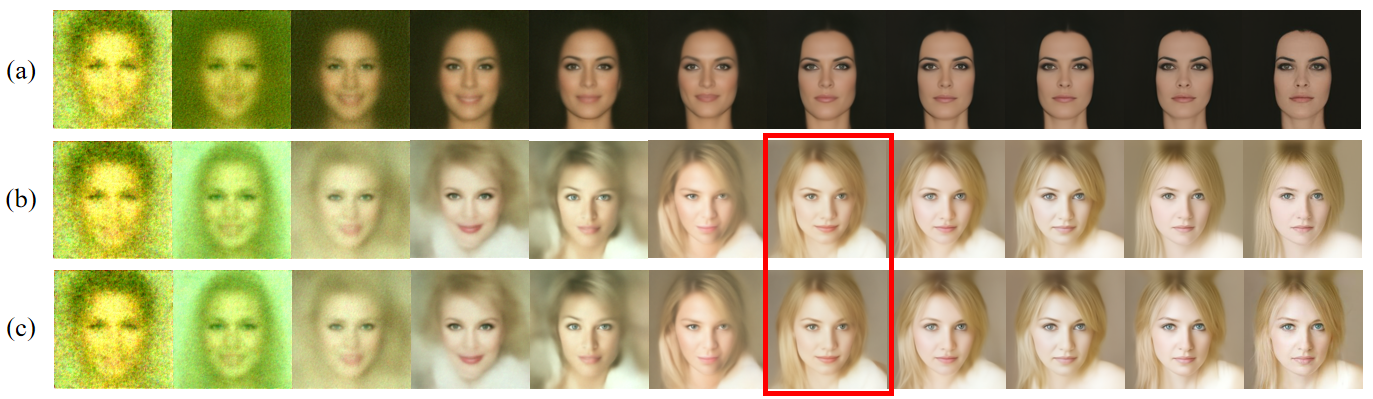}
    \caption{Role of Low-Frequency Noise in Generation.}
    \label{fig1}
\end{figure}
We first examine the role of noise low-frequency components. As shown in Fig.\ref{fig1} (top to bottom), three diffusion sampling strategies with respect to low-frequency noise are compared: complete removal, removal after a certain step and full retention. It can be observed that completely removing low-frequency noise (i.e., setting it to zero)  renders the evolution deterministic, limiting fine-detail refinement and degrading perceptual realism. Partial removal after an initial period preserves global structure but leaves fine details underdeveloped. These results indicate that low-frequency noise plays a crucial role in shaping the global structure, particularly during the early stages of the sampling process.

\begin{figure}[t]
    \centering
    \includegraphics[width=\linewidth]{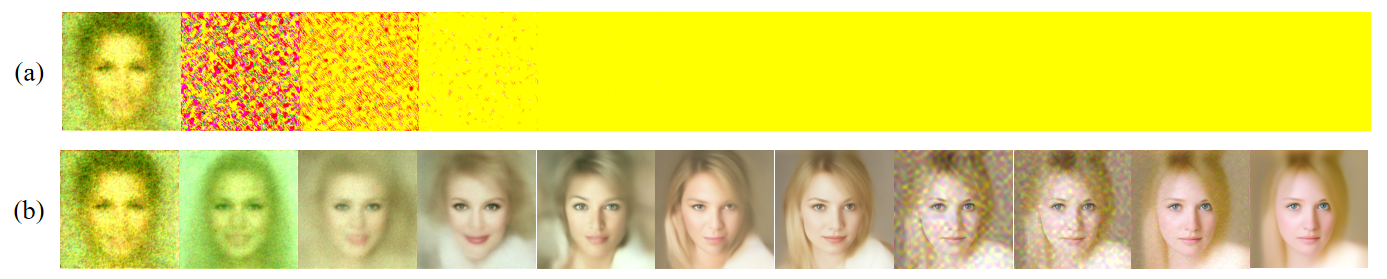}
    \caption{Role of High-Frequency Noise in Generation.}
    \label{fig2}
\end{figure}
Then we examine high-frequency noise components. As shown in Fig.2(a), completely removing high-frequency noise (i.e., setting $\bm{\bar{\epsilon}}_{t-1}^{HF}=0$) causes generated images to degenerate into semantic-free, uniform blobs. Partial removal after structural stabilization introduces patchy artifacts (Fig.2(b)). This demonstrates that high-frequency noise drives early structural evolution and later refines textures; its absence leads to either structural stagnation or excessive smoothing.

Therefore, in generative tasks, the low-frequency components of noise govern the global structure, while the high-frequency components refine details and textures—low frequencies form the skeleton, and high frequencies provide the flesh. Only by properly leveraging noise across frequency bands can generation achieve both structural coherence and visual realism, yielding high-quality images.
\subsection{Role of Injected Noise in Image Restoration}\label{sec3.2}
In image restoration, the goal is to reconstruct a clean image $\bm{x}_0$ consistent with the given measurement $\bm{y}$. The corresponding DDIM sampling update equation is:
\begin{align}
    \bm{x}_{t-1} = \sqrt{\bar{\alpha}_{t-1}} \, {\bm{x}}_{0|t} -\mu\nabla_{\bm{x}}\ell(\bm{x}_{0|t},\bm{y})+ \sqrt{1 - \bar{\alpha}_{t-1}} \, \bar{\bm{\epsilon}}_{t-1},
    \label{eq14}
\end{align}
where $\mu$ is the guidance strength and $\nabla_{\bm{x}}\ell(\bm{x}_{0|t},\bm{y})$ is the data consistency constraint.

By applying frequency-domain decomposition to Eq.(\ref{eq14}), we obtain the decoupled form as follows:
\begin{align}
    \left\{
    \begin{aligned}
    &\bm{x}_{t-1} = \bm{x}_{t-1}^{LF}+\bm{x}_{t-1}^{HF}\\
    &\bm{x}_{t-1}^{LF}=\sqrt{\bar{\alpha}_{t-1}} \, {\bm{x}}_{0|t}^{LF} - \mu\nabla_{\bm{x}}^{LF}\ell(\bm{x}_{0|t},\bm{y})+ \sqrt{1 - \bar{\alpha}_{t-1}} \, \bar{\bm{\epsilon}}_{t-1}^{LF}\\
    &\bm{x}_{t-1}^{HF}=\sqrt{\bar{\alpha}_{t-1}} \, {\bm{x}}_{0|t}^{HF} - \mu\nabla_{\bm{x}}^{HF}\ell(\bm{x}_{0|t},\bm{y}) + \sqrt{1 - \bar{\alpha}_{t-1}} \, \bar{\bm{\epsilon}}_{t-1}^{HF}
    \end{aligned}
    \right..
    \label{eq15}
\end{align}

% In Sec.\ref{sec3.1}, we showed analytically and empirically that full-band Gaussian noise is crucial for high-quality generation. However, this no longer holds for restoration: common degradations (e.g., Gaussian blur or downsampling) act as low-pass filters—strongly attenuating high frequencies while preserving most low-frequency content\cite{wang2025not}.
Let us now examine the roles of different noise components in image restoration tasks. A critical distinction from generation tasks is that in restoration problems—such as deblurring or super-resolution—high-frequency details and textures are significantly attenuated or lost in the degraded observation. Therefore, missing high-frequency information is provided by the diffusion prior in restoration,  which  is precisely the  role of high-frequency component of the injected noise.

\begin{figure}[h]
    \centering
    \includegraphics[width=\linewidth]{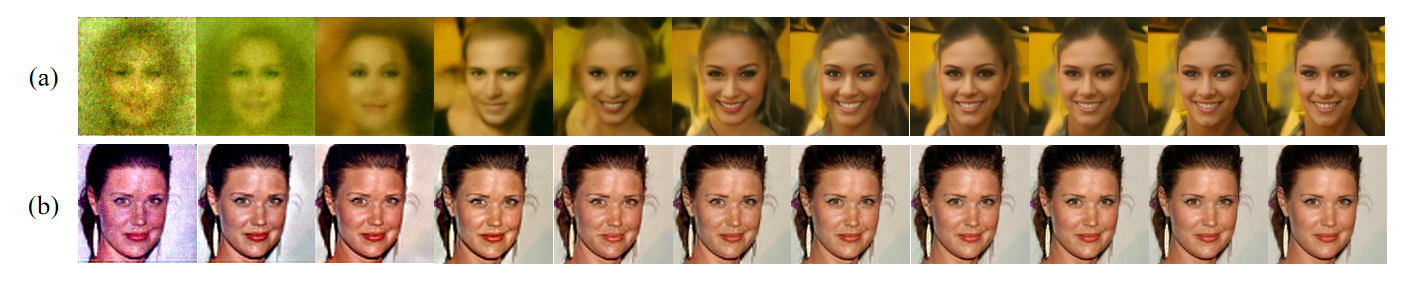}
    \caption{Intermediate processes of generation and restoration.}
    \label{fig3}
\end{figure}

As illustrated in Fig.\ref{fig3}, the intermediate results of the diffusion process of generation Fig.3(a)  and restoration Fig.3(b) are presented. Different from generation, in restoration the initial prediction can accurately recover the global (low-frequency) image structure with the aid of the degraded input, and the errors only at the detail level. This indicates that in restoration, the low-frequency content is strongly constrained by the data-consistency condition.

% Fig.\ref{fig3} further visualizes the intermediate results of DD-NRLG after data consistency correction. Unlike in generative tasks (Fig.3(a)), the initial prediction already accurately recovers the ground-truth macro structure (Fig.3(b)), with errors only present in fine details-indicating that low-frequency components are tightly constrained by data consistency in image restoration.

% However, as discussed earlier, the injected low-frequency noise components introduce unnecessary stochastic perturbations to this deterministic structure, thereby degrading reconstruction fidelity; in contrast, high-frequency noise helps generate rich textural details, which is precisely why current sampling-based PnP methods achieve high perceptual quality but suffer from reduced fidelity.
Consequently, injecting low-frequency noise introduces unnecessary random perturbations, thereby degrading reconstruction fidelity. In contrast, high-frequency noise constructively contributes to synthesizing realistic textures and fine details. This analysis explains why existing sampling-based PnP methods—though capable of high perceptual quality—often suffer from reduced fidelity. Therefore, it is needed to modulate the noise in the diffusion-based restoration.

\subsection{Noise Frequency-Controlled Diffusion Sampling}
% As discussed in Sec.\ref{sec3.2}, low-frequency components of full-band Gaussian noise conflict with the data-consistent structure, degrading fidelity, while high frequencies enhance textural details. To mitigate this, we propose a noise frequency-controlled diffusion sampling (NFCDS) strategy that suppresses low frequencies and retains high frequencies during re-injection, reducing structural interference without compromising texture synthesis.
To achieve a better fidelity-perception trade-off in image restoration, we propose a noise frequency-controlled diffusion sampling (NFCDS) strategy. NFCDS modulates the noise spectrum by suppressing the low frequencies and retaining high ones. This selective filtering reduces structural interference without compromising texture and structures during the diffusion. 

Specifically, at each sampling step, after applying the data consistency constraint and before re-injecting noise, we modulate the noise $\bm{\bar{\epsilon}}_{t-1}$ in the Fourier domain as follows:
\begin{align}
    \bar{\bm{\epsilon}}'_{t-1}=\text{NFCDS}(\bar{\bm{\epsilon}}_{t-1},t)=\bm{\mathcal{F}}^{-1}(\bm{\mathcal{F}}(\bar{\bm{\epsilon}}_{t-1})\odot \bm{\mathcal{M}}(t)),
    \label{eq16}
\end{align}
where $\bm{\mathcal{F}}(\cdot),\bm{\mathcal{F}^{-1}}(\cdot)$ denote the Fourier and inverse Fourier transform; $\odot$ represents element-wise multiplication; and $\bm{\mathcal{M}}(t)$ is a soft-thresholding mask in the Fourier domain that filters out low-frequency components of the noise:
\begin{align}
    \bm{\mathcal{M}}(t)= \frac{1}{1 + \exp\!\left( -\alpha \cdot \big( ||\omega|| - r(t) \big) \right)},
    \label{eq17}
\end{align}
where $\omega=(u,v)$ denotes the frequency-domain coordinates; $r(t)$ controls the low-frequency suppression radius at each sampling step (here we set $r(t)=r_{\text{thresh}}$, a fixed constant requiring hyperparameter tuning); and $\alpha$ governs the mask’s transition steepness near the cutoff frequency—larger $\alpha$ produces a sharper, near hard-threshold response. Finally, Eq.(\ref{eq14}) is given by:
\begin{align}
    \bm{x}_{t-1}=&\sqrt{\bar{\alpha}_{t-1}}{\bm{x}}_{0|t}-\mu\nabla_{\bm{x}}\ell(\bm{x}_{0|t},\bm{y})\nonumber\\
    +&\sqrt{1-\bar{\alpha}_{t-1}}\, \cdot\text{NFCDS}\left(\sqrt{1-\zeta}\,\bm{\epsilon}_{\theta}(\bm{x}_{t},t)+\sqrt{\zeta}\,\bm{\epsilon}_{t}, t\right).
    \label{eq18}
\end{align}
\renewcommand{\thealgorithm}{1} 
    \begin{algorithm}[t]
        \caption{Zero-Shot Image Restoration with Noise Frequency-Controlled Diffusion Sampling}\label{alg1} 
        \begin{algorithmic}[1] 
            \setlength{\itemsep}{0.5em} 
            \Require $\bm{\epsilon}_{\theta}(\cdot,t)$, T, $\bm{y}$, $\bm{A}$, $\left\{\bar{\alpha}_{t}\right\}$, $\mu$
            \State $\mathrm{Initialize}\;\bm{x}_{T}\sim\mathcal{N}(\bm{0},\bm{I})$
            \For{$t\;from\;T\;to\;1$}
                \State $\bm{x}_{0|t}=\frac{1}{\sqrt{\bar{\alpha}_{t}}}(\bm{x_{t}} - \sqrt{1-\bar{\alpha}_{t}} \bm{\epsilon}_{\bm{\theta}}(\bm{x_{t}},t))$
                \State $\hat{\bm{x}}_{0|t}=\bm{x}_{0|t}-\mu\nabla_{\bm{x}}\ell(\bm{x}_{0|t},\bm{y})$
                \State $\bm{\epsilon}_t\sim\mathcal{N}(\bm{0},\bm{I})$
                \State $\bar{\bm{\epsilon}}_{t-1}=\sqrt{1-\zeta}\,\bm{\epsilon}_{\theta}(\bm{x}_{t},t)+\sqrt{\zeta}\,\bm{\epsilon}_{t}$
                \State $\bm{x}_{t-1}=\sqrt{\bar{\alpha}_{t-1}}\hat{\bm{x}}_{0|t}+\sqrt{1-\bar{\alpha}_{t-1}}\text{NFCDS}(\bar{\bm{\epsilon}}_{t-1}, t)$   
            \EndFor
            \State \Return $\bm{x}_{0}$
        \end{algorithmic}
    \end{algorithm}

% Through this operation, we actively suppress the most detrimental low-frequency perturbations while preserving sampling stochasticity, thereby establishing a frequency-domain control mechanism. This mechanism is applicable to any zero-shot image restoration method that relies on noise re-injection to achieve high perceptual quality, and is summarized in Algorithm \ref{alg1}.
% This operation actively suppresses the most detrimental low-frequency perturbations while preserving sampling stochasticity, establishing a frequency-domain control mechanism. It applies to any zero-shot image restoration method that uses noise re-injection for high perceptual quality and is summarized in Algorithm \ref{alg1}.
This frequency-domain control mechanism effectively suppresses detrimental low-frequency perturbations and reduces the number of required sampling steps by making early iterations more effective, thereby accelerating the process. It is broadly applicable to any zero-shot image restoration method that achieves high perceptual quality via noise re-injection, as summarized in Algorithm\ref{alg1}.
% This frequency-domain control mechanism effectively suppresses detrimental low-frequency perturbations while eliminating early ineffective iterations, significantly reducing sampling steps and accelerating the process. It is broadly applicable to any zero-shot image restoration method that achieves high perceptual quality via noise re-injection, as summarized in Algorithm\ref{alg1}.

\section{Experiment}\label{sec4}
% In this section, we conduct experiments on two image restoration tasks, super-resolution and Gaussian denoising, to evaluate the improvement of our method over existing diffusion-based zero-shot restoration approaches. We primarily adopt DD-NRLG as the baseline framework, using the same parameter settings as in the original algorithm to eliminate the influence of additional hyperparameter tuning. Furthermore, we perform partial experiments on other methods such as DiffPIR and DDPG to validate the generalizability of the proposed NFCDS. All experiments employ a publicly available pre-trained model trained on the CelebA-HQ (256×256) benchmark dataset using the DDPM noise schedule.
We evaluate our method on two zero-shot image restoration tasks—super-resolution and Gaussian denoising—against existing diffusion-based approaches. Using DD-NRLG as the primary baseline with its original parameters to avoid hyperparameter tuning effects, we also compare with mainstream methods including DDRM\cite{ddrm}, DDNM\cite{ddnm}, and DDPG\cite{ddpg}. The values of $r_{\text{thresh}}$ and $\alpha$ for NFCDS applied to DD-NRLG are given in Table \ref{table1}. All experiments use the publicly available DDPM model pre-trained on CelebA-HQ (256×256).
\begin{table}[!h]
    \centering
    \renewcommand{\arraystretch}{1}
    \setlength{\tabcolsep}{5pt}
    \caption{The selection of parameters.}
    \resizebox{0.9\linewidth}{!}{
    \begin{tabular}{
        l 
        c c 
        c c c 
        }
        \toprule
        \textbf{DD-NRLG}
        & \multicolumn{2}{c}{SR×4} 
        & \multicolumn{3}{c}{Denosing}
        \\
        \cmidrule(lr){2-3} \cmidrule(lr){4-6} 
        \textbf{Parameters} 
        & $\sigma_{\bm{y}}=0$ & $\sigma_{\bm{y}}=0.05$ 
        & $\sigma_{\bm{y}}=0.1$ & $\sigma_{\bm{y}}=0.25$ & $\sigma_{\bm{y}}=0.5$\\
        \midrule
        $r_{\text{thresh}}$      & 35 & 25 & 64 & 35 & 25   \\
        $\alpha$      & 5 & 5 & 5 & 5 & 5  \\
        \bottomrule
    \end{tabular}
    }
    \label{table1}
\end{table}
\begin{table*}[t]
    \centering
    \renewcommand{\arraystretch}{1.3}
    \setlength{\tabcolsep}{5pt}
    \caption{Quantitative results of two tasks on the CelebA-HQ dataset.}
    \resizebox{0.9\textwidth}{!}{
    \begin{tabular}{
        l c
        c c c 
        c c c 
        c c c 
        c c c 
        c c c
        }
        \toprule
        \textbf{CelebA-HQ}&
        & \multicolumn{3}{c}{SR×4($\sigma_{\bm{y}}=0.$)} 
        & \multicolumn{3}{c}{SR×4($\sigma_{\bm{y}}=0.05$)}
        & \multicolumn{3}{c}{Denoising($\sigma_{\bm{y}}=0.1$)}
        & \multicolumn{3}{c}{Denoising($\sigma_{\bm{y}}=0.25$)}
        & \multicolumn{3}{c}{Denoising($\sigma_{\bm{y}}=0.5$)} \\
        \cmidrule(lr){3-5} \cmidrule(lr){6-8} \cmidrule(lr){9-11}
        \cmidrule(lr){12-14} \cmidrule(lr){15-17}
        \textbf{Method} & NFE 
        & PSNR $\uparrow$ & LPIPS $\downarrow$ & SSIM $\uparrow$
        & PSNR $\uparrow$ & LPIPS $\downarrow$ & SSIM $\uparrow$
        & PSNR $\uparrow$ & LPIPS $\downarrow$ & SSIM $\uparrow$
        & PSNR $\uparrow$ & LPIPS $\downarrow$ & SSIM $\uparrow$ 
        & PSNR $\uparrow$ & LPIPS $\downarrow$ & SSIM $\uparrow$\\
        \midrule
        DDRM   & 100 & 31.98 & 0.057 & \underline{0.8848} & 29.53 & \textbf{0.085} & 0.8313 & 33.47 & 0.054 & 0.9090 & 30.39 & 0.083 & 0.8613 & 28.01 & 0.124 & 0.8135 \\ 
        % DPS\cite{dps}      & 31.03 & 0.059 & 0.8857 & 28.59 & 0.084 & 0.8017 & 26.62 & \textbf{0.092} & 0.7545 & 28.33 & 0.190 & 0.7917 & 25.96 & 0.234 & 0.7357\\
        DDNM    & 100 & 31.96 & \textbf{0.049} & 0.8811 & - & - & - & \underline{34.01} & 0.053 & \textbf{0.9147} & \underline{30.79} & \underline{0.079} & \underline{0.8631} & 28.19 & \underline{0.099} & 0.8147 \\
        DDPG    & 100 & 31.94 & \underline{0.051} & 0.8818 & 29.57 & 0.106 & \underline{0.8329} & 33.42 & 0.055 & 0.8800 & 28.43 & 0.168 & 0.8186 & 25.47 & 0.179 & 0.7651 \\
        \midrule
        DD-NRLG      & 100 & \underline{32.07} & \textbf{0.049} & 0.8834 & 29.37 & \underline{0.097} & 0.8237 & 33.51 & \textbf{0.046} & 0.9037 & 30.43 & \underline{0.079} & 0.8549 & \underline{28.20} & 0.121 & 0.8162 \\
        DD-NRLG   & \textbf{50}  & 31.86 & 0.062 & 0.8793 & 29.15 & 0.114 & 0.8172 & 32.95 & 0.059 & 0.8741 & 29.72 & 0.089 & 0.8098 & 27.34 & 0.133 & 0.7693 \\
        \midrule
        DD-NRLG + NFCDS    & \textbf{50}  & \textbf{32.12} & \underline{0.051} & \textbf{0.8912} & \textbf{29.70} & \textbf{0.085} & \textbf{0.8387} & \textbf{34.10} & \underline{0.048} & \underline{0.9101} & \textbf{31.00} & \textbf{0.072} & \textbf{0.8634} & \textbf{28.60} & \textbf{0.096} & \textbf{0.8215} \\
        \bottomrule
    \end{tabular}
    }
    \label{table2}
\end{table*}
\begin{figure*}[t]
    \centering
    \includegraphics[width=0.8\linewidth]{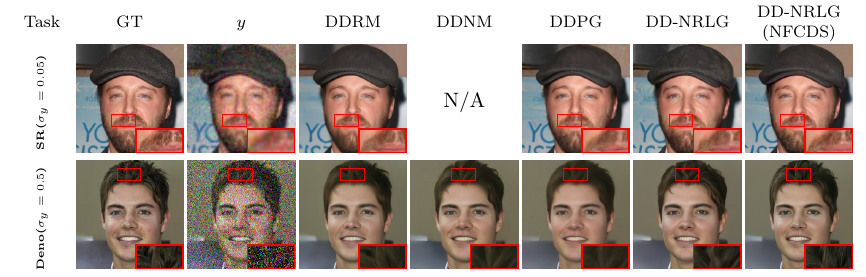}
    \caption{Visual comparison of results (N/A: method is unable to handle the task).}
    \label{fig4}
\end{figure*}

Super-resolution uses bicubic downsampling with scale factor 4, under both noise-free and noisy $(\sigma_{\bm{y}}=0.05)$ conditions. Gaussian denoising is evaluated at noise levels $\{0.1, 0.25, 0.5\}$. Performance is measured by PSNR, LPIPS, and SSIM. 

Quantitative results are reported in Table \ref{table2}. Our method achieves the best or second-best performance in all settings, consistently improving PSNR and SSIM over DD-NRLG while preserving strong LPIPS scores. In noisy super-resolution $\sigma_{\bm{y}}=0.05$, the gains in PSNR and SSIM are particularly pronounced, with LPIPS achieving the best value. For denoising under strong noise $\sigma_{\bm{y}}=0.5$, integrating NFCDS into DD-NRLG yields a clear improvement: PSNR increases by 0.44 dB and LPIPS decreases by nearly 0.02. 

Moreover, we observe that incorporating NFCDS reduces ineffective iterations and significantly lowers the required number of sampling steps, with detailed timing improvements reported in Table \ref{table3}.
\begin{table}[h]
    \centering
    \renewcommand{\arraystretch}{0.5}
    \setlength{\tabcolsep}{10pt} % 略微减小以适应更多列
    \caption{Inference time comparison.}
    \resizebox{0.5\textwidth}{!}{
    \begin{tabular}{@{\extracolsep\fill}lccccc@{\extracolsep\fill}}
    \toprule
     & NFCDS & NFE & SR & Deno\\ 
    \midrule
    DD-NRLG & \ding{55} & 100 & 8.02s & 7.92s  \\ 
    DD-NRLG & \ding{51} & 50 & 4.16s & 4.12s \\ 
    \midrule
    Comparison & - & - & {\textbf{$\downarrow$3.86s}} & {\textbf{$\downarrow$3.80s}}\\ 
    \bottomrule
    \end{tabular}
    }
    \label{table3}
\end{table}

The visual comparison as shown in Fig.\ref{fig4} demonstrates that the restoration results, after incorporating the NFCDS mechanism, exhibit more realistic and detailed textures, such as hair and beards. In the super-resolution task, the background is free of the artifacts observed in DD-NRLG.

Furthermore, we conduct super-resolution experiments on both DiffPIR and DDPG to verify the generalizability of NFCDS; quantitative results in Table \ref{table4} show that metrics improve significantly after applying NFCDS.
\begin{table}[t]
    \centering
    \renewcommand{\arraystretch}{1.3}
    \setlength{\tabcolsep}{5pt}
    \caption{Generalization of NFCDS on DiffPIR and DDPG}
    \resizebox{0.95\linewidth}{!}{
    \begin{tabular}{
        l 
        c c c 
        c c c 
        }
        \toprule
        \textbf{CelebA-HQ}
        & \multicolumn{3}{c}{SR×4($\sigma_{\bm{y}}=0$)} 
        & \multicolumn{3}{c}{SR×4($\sigma_{\bm{y}}=0.05$)}
        \\
        \cmidrule(lr){2-4} \cmidrule(lr){5-7} 
        \textbf{Method} 
        & PSNR $\uparrow$ & LPIPS $\downarrow$ & SSIM $\uparrow$
        & PSNR $\uparrow$ & LPIPS $\downarrow$ & SSIM $\uparrow$\\
        \midrule
        DiffPIR      & 29.83 & 0.085 & 0.8352 & 26.81 & 0.096 & 0.7637  \\
        DiffPIR + NFCDS      & 30.02 & 0.090 & 0.8458 & 27.13 & 0.097 & 0.7714 \\
        \midrule
        DDPG      & 31.80 & 0.051 & 0.8829 & 29.57 & 0.106 & 0.8329  \\
        DDPG + NFCDS      & 32.12 & 0.062 & 0.8914 & 30.01 & 0.100 & 0.8412 \\
        \bottomrule
    \end{tabular}
    }
    \label{table4}
\end{table}

\section{Conclusion}
% This work addresses the common issue in sampling-based Plug-and-Play methods where high perceptual quality comes at the cost of reduced fidelity. We present the first frequency-domain analysis that disentangles the roles of different noise frequency components in generative and restoration tasks, revealing the negative impact of low-frequency noise on image recovery. Based on this insight, we propose a Noise Frequency-Domain Control (NFDC) mechanism that suppresses low-frequency components of the injected noise in the Fourier domain, thereby significantly improving reconstruction fidelity while preserving visual quality.
This work addresses the well-known trade-off in sampling-based Plug-and-Play methods, where high perceptual quality is often achieved at the expense of reduced reconstruction fidelity. We analyze the roles of different noise components from a frequency-domain perspective in both generative and restoration tasks, revealing the detrimental impact of low-frequency noise on reconstruction fidelity. Building on this insight, we propose a Noise Frequency-Controlled Diffusion Sampling (NFCDS) strategy that suppresses the low-frequency components of the injected noise in the Fourier domain. Notably, NFCDS requires no additional training and can be seamlessly integrated into existing frameworks, significantly improving reconstruction fidelity while preserving high visual quality.

\bibliographystyle{IEEEbib}
\bibliography{strings,refs}

\end{document}